\documentclass[conference]{IEEEtran}
\IEEEoverridecommandlockouts
\usepackage{cite}
\usepackage{amsmath,amssymb,amsfonts}
\usepackage{algorithmic}
\usepackage{graphicx}
\usepackage{textcomp}
\usepackage{xcolor}
\def\BibTeX{{\rm B\kern-.05em{\sc i\kern-.025em b}\kern-.08em
    T\kern-.1667em\lower.7ex\hbox{E}\kern-.125emX}}

\usepackage{framed,multirow}

\usepackage{amssymb}
\usepackage{latexsym}
\usepackage{subfigure}
\usepackage{amsmath}
\usepackage{amssymb}
\usepackage{booktabs}
\usepackage{commath}
\usepackage[pagebackref,breaklinks,colorlinks]{hyperref}

\usepackage{array}

\newcolumntype{C}[1]{>{\centering\arraybackslash}p{#1}}
\usepackage{url}
\usepackage{xcolor}
\definecolor{newcolor}{rgb}{.8,.349,.1}

\begin{document}

\title{Patterns of Vehicle Lights: Addressing Complexities in Curation and Annotation of Camera-Based Vehicle Light Datasets and Metrics}

\author{Ross Greer, Akshay Gopalkrishnan, Maitrayee Keskar, Mohan M. Trivedi\thanks{Note: This preprint is in the process of submission for consideration to Pattern Recognition Letters.}}



\maketitle





\begin{abstract}
This paper explores the representation of vehicle lights in computer vision and its implications for various tasks in the field of autonomous driving. Different specifications for representing vehicle lights, including bounding boxes, center points, corner points, and segmentation masks, are discussed in terms of their strengths and weaknesses. Three important tasks in autonomous driving that can benefit from vehicle light detection are identified: nighttime vehicle detection, 3D vehicle orientation estimation, and dynamic trajectory cues. Each task may require a different representation of the light. The challenges of collecting and annotating large datasets for training data-driven models are also addressed, leading to introduction of the LISA Vehicle Lights Dataset and associated Light Visibility Model, which provides light annotations specifically designed for downstream applications in vehicle detection, intent and trajectory prediction, and safe path planning. A comparison of existing vehicle light datasets is provided, highlighting the unique features and limitations of each dataset. Overall, this paper provides insights into the representation of vehicle lights and the importance of accurate annotations for training effective detection models in autonomous driving applications. Our dataset and model are made available at https://cvrr.ucsd.edu/vehicle-lights-dataset
\end{abstract}





\section{Introduction: Why Detect Lights?}
%
%
%
%
Computer vision, defined broadly, is a term which describes methods for obtaining useful information from images. A typical task within computer vision is \textit{object detection}, in which the derived information is the location and concept of objects contained in an image; helpful information when making decisions related to path-planning \cite{bajkowski2023comparing}. However, depending on the particular engineering application for which this information is intended, specification of \textit{location} and \textit{concept} may vary. Past research and associated datasets have painted detected lights as a pixel-wise segmentation mask over an image \cite{rapson2018reducing}, a center coordinate of a light region with a bounding box \cite{o2010vehicle}, and collection of non-rectangular corner points which approximately bound the light \cite{song2019apollocar3d}. 

Through this research, we explore this question of information representation by presenting a variety of specifications and their strengths and weaknesses towards the task of vehicle light detection. Additionally, as data-driven methods of detection have become a dominant approach in the field, we present associated challenges in collecting, curating, and annotating large datasets fit for this task. Finally, we introduce a dataset which contains a representation of light information designed specifically for downstream applications in vehicle detection, intent and trajectory prediction, and safe path planning.

\begin{figure*}
    \centering
    \includegraphics[width=\textwidth]{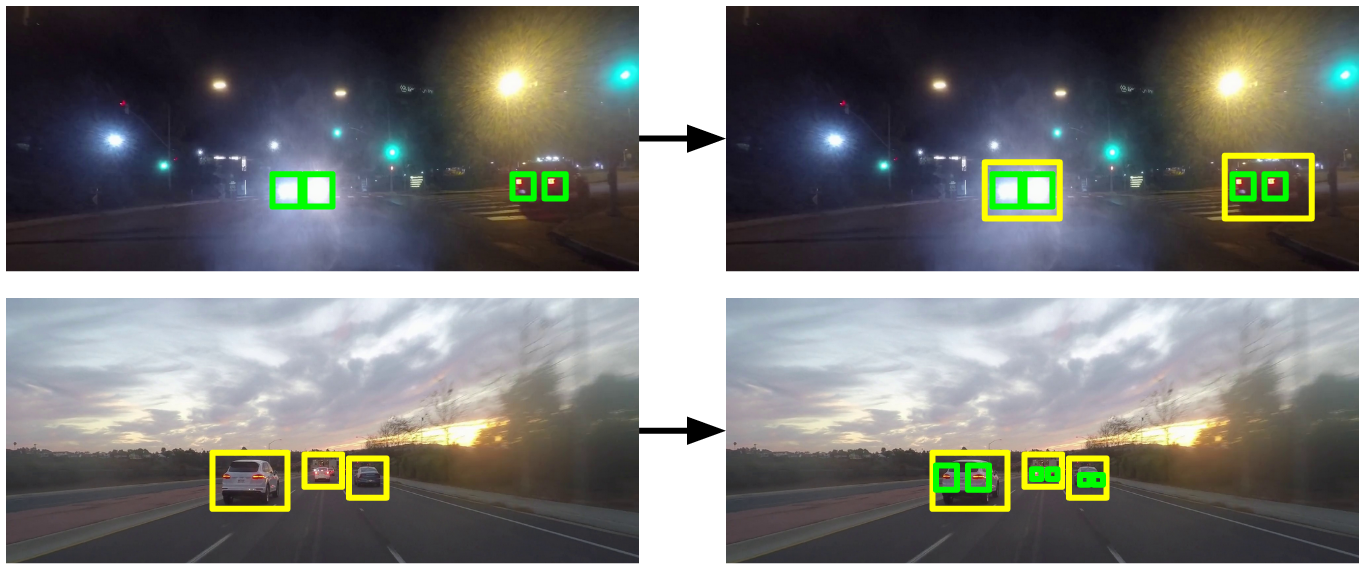}
    \caption{In both the top and bottom scenarios, detection of vehicle lights plays an important role in planning a safe vehicle trajectory. In the top row, the vehicle lights can be seen even though the body of the left vehicle itself is not visible, allowing a model to infer the vehicle's presence. In the second row, the vehicle's taillights may give indication of braking or an intention to change lanes, adjusting the plans of the ego vehicle. In these cases, the order of detections (lights-then-vehicle, or vehicle-then-lights) depends on both the task and environment.}
    \label{fig:order}
\end{figure*}

Though end-to-end deep learning approaches often present as the simplest approach to problems in the intelligent and autonomous vehicles domain, it is often difficult to gather sufficiently representative data to train these models. Feature extraction and engineering provide an alternative to end-to-end methodologies, where features known to be meaningful towards the task are explicitly extracted \cite{narote2018review} \cite{sun2004object} rather than implicitly learned, providing a layer of explainability to any system. So, what existing problems in autonomous driving can vehicle light detection play an important step in addressing? Described here and illustrated in Figure \ref{fig:order}, we present example tasks made easier with vehicle light information; interestingly, each may benefit from a different representation of the light. 

In this paper, we formalize and discuss benefits, challenges, and tradeoffs in representing vehicle lights, and introduce the LISA Vehicle Lights Dataset, an extended annotation of the ApolloCar3D dataset \cite{song2019apollocar3d}.  

\section{Tasks and Representations: Example Cases}

\subsection{Nighttime Vehicle Detection}

Vehicle detection in low-light conditions is challenging for systems not equipped with LiDAR or radar sensors, as the vehicle body may not be visible to a camera. However, a vehicle's visible headlights or taillights provide a tool for inferring the body of the vehicle. Knowing the vehicle's precise location is important for autonomous applications in obstacle avoidance and path planning, ADAS emergency braking, and even infrastructure-based traffic monitoring. For these tasks, the choice in representation of the vehicle light itself is less crucial than the ability of this representation to facilitate accurate localization and bounding of the vehicle itself; the relationship between the centroids of the lights and the size of the lights may provide information towards the orientation and proximity of the observed vehicle. 

The low light conditions can introduce further challenges in distinguishing between vehicle lights and other luminous objects, such as streetlights and signs, so any algorithm for detecting a vehicle light must not be so general that any scene light object is detected. 

\subsection{Vehicle Identification and Tracking}

For the purposes of identifying (or re-identifying) a particular vehicle instance, a vehicle's lights can contribute to a vehicle's unique signature \cite{zhu2023dual}, helpful in distinguishing two vehicles that are otherwise very similar (e.g. two white minivans of similar model, but with tail light damage on one vehicle), and even useful as detections to be filtered when localizing the license plate \cite{wang2008fuzzy}. For purposes of both short and long-term tracking of vehicles, being able to identify the same vehicle between frames (or in some cases, across scenes) is an important task, and when these systems are designed end-to-end, the vehicle lights can be a confounding factor in model performance \cite{bashir2019vr}.

\subsection{3D Vehicle Orientation}

Understanding the 3D world location of vehicles in a driving scene is important for safe path planning, and beyond knowing a surround vehicle's location, the orientation provides information that is highly informative to predicting the direction of a vehicle's future motion. Cameras are typically not as accurate as LiDAR or radar for this task, but using cameras to detect centers or corners of lights may be useful for this task, as there is a known symmetry to vehicle layout which may be leveraged to determine which way the car is heading based on the relative positions of the lights in the image. 

\subsection{Dynamic Trajectory Patterns}

A vehicle's lights are a means by which a driver can communicate to surrounding vehicles, providing signals of their intention to turn, change lanes, or current braking. These cues are temporal in nature, where some pattern of blinking or sustained light may indicate a turn signal or brake signal. These cues are visual in nature, and to effectively extract these signals, a camera-based system must be able to identify the region of the image containing the light and then analyze this region, making a bounding box representation most effective to this task. With these cues available, it becomes possible to enhance models that perform vehicle maneuver classification and trajectory predictions \cite{deo2018would, messaoud2021trajectory, deo2020trajectory, greer2021trajectory}. \\

\section{Representation: What is a Light?} 

\begin{figure}
    \centering
    \begin{subfigure}
      \centering
      \includegraphics[width=.4\linewidth]{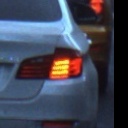}  
    \end{subfigure}
    \begin{subfigure}
      \centering
      \includegraphics[width=.4\linewidth]{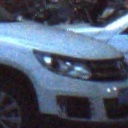}  
    \end{subfigure}
    
    \begin{subfigure}
      \centering
      \includegraphics[width=.4\linewidth]{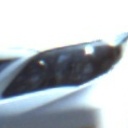}  
    \end{subfigure}
    \begin{subfigure}
      \centering
      \includegraphics[width=.4\linewidth]{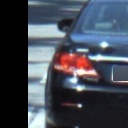}  
    \end{subfigure}
    \caption{Examples of vehicle lights under normal visibility conditions.}
    \label{fig:lights_normal}
\end{figure}

\begin{figure}
    \centering
    \begin{subfigure}
      \centering
      \includegraphics[width=.4\linewidth]{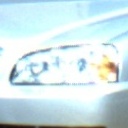}  
    \end{subfigure}
    \begin{subfigure}
      \centering
      \includegraphics[width=.4\linewidth]{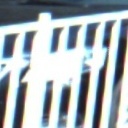}  
    \end{subfigure}
    
    \begin{subfigure}
      \centering
      \includegraphics[width=.4\linewidth]{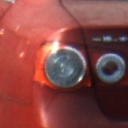}  
    \end{subfigure}
    \begin{subfigure}
      \centering
      \includegraphics[width=.4\linewidth]{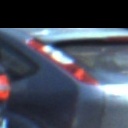}  
    \end{subfigure}
    \caption{Examples of vehicle lights under challenging conditions such as irregular light shapes, vehicle light occlusion, lighting conditions, and minimal vehicle light-color contrast on vehicle.}
    \label{fig:lights_challenging}
\end{figure}

Vehicle lights refer to headlights (lights found in the front of the vehicle) and taillights (lights found in the back of the vehicle). Headlights are typically used for illumination, while taillights typically indicate the vehicle's presence and direction of travel to other vehicles, consisting of a red brake light, a red turn signal light, and a white or amber backup light. The exact location and size of these assemblies of lights vary between automakers. Regulators require that taillights be red and visible from 500 feet, headlights be white or amber and visible from 1,000 feet. Turn signals are sometimes integrated into headlight assemblies, though this is not true of all vehicles. 

\subsection{Lights as Boxes}

\begin{figure}
    \centering
    \begin{subfigure}
      \centering
      \includegraphics[width=.45\linewidth]{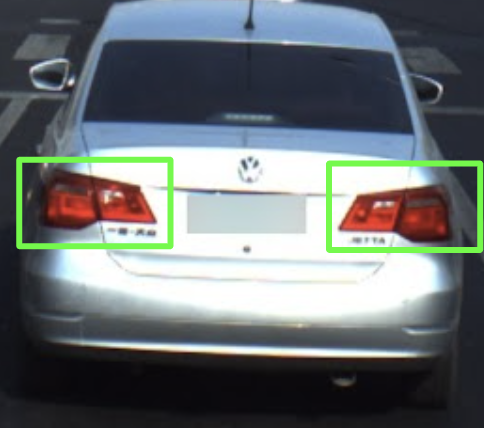} 
    \end{subfigure}
    \begin{subfigure}
      \centering
      \includegraphics[width=.45\linewidth]{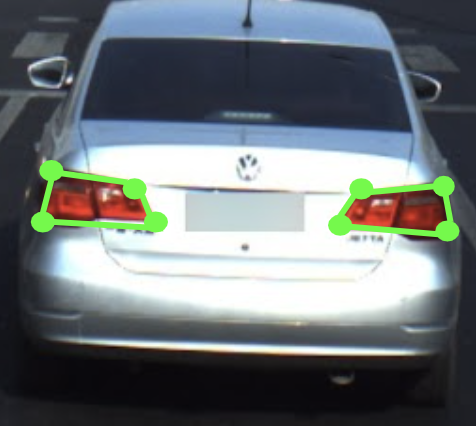}  
    \end{subfigure}
    
    \caption{The left image offers an example of what a vehicle light box annotation would look like for the dataset Jeon et al. \cite{jeon2022deep} uses for their taillight detection approach. The right image visualizes how a vehicle light can be represented by using its four corners, which is used in the ApolloCar3D Dataset \cite{song2019apollocar3d}. Depending on the intended application for the detected lights, such as turn signal cue extraction or 3D pose estimation, one representation may be preferred to another.}. 
    \label{fig:light_comparison}
\end{figure}

Many 2D object detection models predict a bounding box around the object of interest; there are a few ways to encode a 2D box with four pieces of information, and the some typical encodings include (1) a 2D origin point, a height, and a width, or (2) two 2D points, and different encodings may be easier learned by different models.
This first method of light representation is the specification of a rectangular box: 
\begin{equation}
    l = \{ (x_\text{top left}, y_\text{top left}), (x_\text{bottom right}, y_\text{bottom right})\},
\end{equation}
and a dual representation of a corner with a width and height specified could be formulated as: 
\begin{equation}
    l = \{ (x_\text{top left}, y_\text{top left}), w, h\},
\end{equation}
where $w$ and $h$ are the width and height. 
In some computer vision applications, the defined ground truth labels may limit the capabilities of a model \cite{akers2023simulated}. In this case, because lights themselves are not rectangular, there is inherent ambiguity in the appropriate box for a light. The box can entirely and tightly contain the light, but doing so may contain extraneous visual information. To perform taillight detection, Jeon et al. \cite{jeon2022deep} represent lights as boxes by using the automatic labeling tool MELT \cite{rapson2018reducing} on datasets from Sungkyunkwan University (SKKU). Figure \ref{fig:light_comparison} highlights what the taillight box annotation style of this dataset would look like on an example vehicle, highlighting that such box labels of vehicle lights have significantly less strict bounds than other annotation methods which may be better form-fitting to the shape of the light. As a result, detecting vehicle lights with bounding boxes tends to be (in some regards) simpler since the ground truth box labels provide much more room for error. While these box annotations may not perfectly fit the vehicle light, they can still be sufficient for downstream tasks such as vehicle light state classification where the outside visual information may be acceptable as long as the light itself is fully contained. 

\subsection{Lights as Centers}

Zhou et al. \cite{zhou2019objects} take an approach to object detection which departs from 2D bounding boxes, instead viewing an object as collection of interconnected keypoints, developing CenterNet, which models an object by a single center point (presumed to be akin to the center of a traditional 2D bounding box). Using this center point, the detector can regress to other properties like the object corners, orientation, or even pose.

Drawing from these ideas, analogous to human pose estimation where a person's estimated skeleton is a collection of joint keypoints, the vehicle light can also be conceptualized as a single center point, a ``joint" of a vehicle's body:
\begin{equation}
    l = \{ (x, y) \}
\end{equation}

This representation is appropriate for settings where the visual region is of less importance than the light's location (for example, situations where the geometric relationship between the pairs of headlights or taillights might be useful in estimating vehicle orientation). 

\subsection{Lights as Corners}

In further extension of the idea of lights and vehicle components as a collection of keypoints, we can consider the light to be a collection of bounding ``corners":

\begin{equation}
    l = \{ (x_1, y_1), (x_2, y_2), ..., (x_n, y_n) \},
\end{equation}
where $n$ is a constant. These are not strictly corners, but moreso points that form an approximate contour around the light region. This has the benefit of adding a light's span to the previous center representation, without the non-fitting rigidity of the bounding box representation. 

One benefit of the prediction of keypoint approaches (e.g. Law \& Deng's CornerNet \cite{law2018cornernet}) is that the boundary features of the object may be easier to identify as singular points, rather than learning the span of an object through anchor-box learning in methods like Faster RCNN \cite{girshick2015fast} or YOLO \cite{redmon2016you}. A further example of this style of object detection, Dörr et al. \cite{dorr2022tetrapacknet} base their approach off of CornerNet with TetraPackNet, but instead of predicting a top-left corner and bottom-right corner, an object is represented as four annotated vertices. In this case, fixed landmark points of interest are being estimated (rather than a broad parameterization of a bounding box as height, width, and origin, for which many possible encodings exist which define the same box). Zhao et al. \cite{zhao2022corner} refine CenterNet's ability to differentiate corners of objects by adding box-edge length constraints, introducing the lower-cost CenternessNet. An example of a light as corners is shown on right in Figure \ref{fig:light_comparison}. 

\subsection{Lights as Masks}

The lights of a vehicle can be identified on a pixel-wise level, marking any pixels which make up the light as a segmentation mask:

\begin{equation}
    l = \{ (x_1, y_1), (x_2, y_2), ..., (x_p, y_p) \},
\end{equation}
where the value of pixels in a mask, $p$, can vary between lights. This approach is the most specific approach to defining the light, but deprives the ability to find associated keypoint pairs between the lights for the purposes of finding geometric relationships between twin taillights or headlights of a vehicle. Further, this approach requires postprocessing to extract a cropped region of interest around the light for use in downstream models, and depending on how the mask is estimated, this can lead to many extraneous points being included in the estimated region. 

\subsection{Translating Between Representations}

Accepting some loss of specificity, it is possible to convert between some representations of a light. This may be useful if the dataset useful to a task has been annotated with a representation nonstandard to the intended model, or for computing metrics for comparison between different datasets and models. The center-based representation is most restrictive, as there is no means of inferring the span of the light from the center alone. Here, we suggest some methods of conversion between representations: 
\begin{itemize}
    \item The center of a light can be found from the bounding box, corners, or mask by taking the centroid of the given coordinates.
    \item The bounding box of a light can be found from corners or a mask by taking the top left corner to be $(\min(x), \min(y))$ and the bottom right corner to be $(\max(x), \max(y))$, taken over all corners or all light-class pixels of the mask. Note that it is important in this case that the mask be a joined and filtered connected component (i.e. no fragments or noisy pixels in the representation). 
\end{itemize}

\subsection{Metrics for Evaluating Light Detection Performance}


For a 2D box representation or mask representation, Intersection over Union (IoU) can quantify how strongly an estimated box fits to an annotation. A threshold on IoU values can be selected, which allows classification of a guessed detection into the usually categories of ``hit" (true positive), ``miss" (false positive), and additionally counting the object instances which the detector failed to find (false negatives). These rates can be combined to form the common mean average precision metric, a metric usually characterized by the threshold applied to the object detector. Additionally, when it comes to road objects, it is common to stratify such results by attributes such as an object's ``size" or ``difficulty", as a means of quantifying detector performance in balance with the complexity of the analyzed scene. 

Likewise, when using a center-based or corner-based representation, an L2-distance measurement could be an appropriate metric for model performance on object instances in this setting. However, it is important to note that for objects further from the camera, the same L2 pixel distance may have a drastically different real-world distance. For this reason, scaling the distance relative to some other attribute (such as detected vehicle width) may help to tune the metric in a way that more accurately reflects the performance of the model. Similarly to the above paragraph, these distance errors can be thresholded to provide a criteria for a ``hit" or ``miss", allowing standard aggregate metrics such as precision and recall to be computed for the light detection model.

\section{Vehicle Light Datasets for Learning Models of Light Patterns}

Tasks related to vehicle lights cover a large span of interest, from measurement and validation of beam luminosity by test and validation engineers, to control conditions for emergency braking ADAS in deployed vehicles. Because of this, there are a variety of features of the light assembly and operation that may be important for a particular task, including but not limited to:
\begin{itemize}
    \item Light Position (Headlight, Tailight, Left, Right)
    \item Light Location (specification of center, corners, box, or mask for detection)
    \item Light Color
    \item Light State (On, Off, High Beam, Low Beam)
    \item Light Signal (Turn Signal, Brake Light, Emergency Lights)
\end{itemize}

In this work, we focus on datasets which are useful toward the task of vehicle light \textit{detection}, centered on the first two static elements of the above list. As deep learning approaches become the common technique for vehicle light detection, there is a need for substantial public vehicle light datasets that can be used to train and evaluate such detection models. However, despite the rapid growth in autonomous driving datasets \cite{bogdoll2023impact}, there are still gaps in vehicle light detection datasets with direct association to a vehicle instance, which is an important relation for the three tasks defined in the introduction. Unlike some autonomous driving data that can be collected and automatically processed from sensor data \cite{satzoda2014drive}, a vehicle light detection dataset is time-consuming to curate since it requires manually annotating vehicle lights in traffic scenes, including time-variant attributes. To expand such datasets, an automatic labeling tool such as 3D BAT \cite{zimmer20193d} can be used on large autonomous vehicle datasets (KITTI \cite{geiger2012we}, A9-Dataset \cite{cress2022a9}, nuScenes \cite{caesar2020nuscenes}, etc.) to label vehicle lights. Rapson et al. \cite{rapson2019performance} construct the Vehicle Light Datasets, which contains manual segmentation labels of vehicle lights from a variety of datasets such as KITTI \cite{geiger2012we}, Berkeley Drive \cite{xu2017end}, and Cityscapes \cite{cordts2016cityscapes}. These segmentation labels of vehicle lights are not necessarily helpful for detection methods that treat vehicle lights as bounding boxes, requiring careful conversion of these segmentation labels to bounding boxes through clustering algorithms and bounding heuristics which may produce inaccurate ground truths. Hsu et al. \cite{hsu2017learning} introduce the Vehicle Rear Signal dataset, which contains consecutive video frames of the back of a vehicle to recognize taillight signals. However, this dataset is not useful for vehicle light \textit{detection} as there is no data related to the location and size of the taillights. Vancea et al. \cite{vancea2017vehicle} note that to train their vehicle light detection model they manually annotate 677 images from the CompCars dataset \cite{yang2015large}. Such a dataset size may be insufficient to train a robust vehicle light detector. The Apollo-Car3D dataset \cite{song2019apollocar3d} contains 5,277 driving images and over 60K car instances, where each vehicle is provided with annotations of labeled keypoints of various parts of the car, including the vehicle lights. This dataset offers the largest amount of vehicle light data but needs to be carefully filtered to remove irrelevant data related to vehicle lights and only include vehicle instances with a visible vehicle light. By processing this dataset to contain only visible light data from the detected vehicles, we create the LISA Vehicle Lights dataset.

\section{Introducing LISA Vehicle Lights Dataset}

In this section, we introduce a sequence of filtering and annotation extensions to the AppoloCar3D dataset, meant to facilitate development of cascaded models which leverage information of vehicle detection instances to extract further information on the vehicle's lights. Starting from given center coordinate annotations of vehicle lights from the ApolloCar3D dataset \cite{song2019apollocar3d}, we create and share a dataset of cropped vehicle light images which contain coordinates and visibility attributes of light assembly corners for each vehicle of interest. 

\subsection{Dataset Features}

\begin{table*}[hbt!]
\caption{Comparison of different public datasets used to train vehicle light detection models. We note that Vancea et al.    \cite{vancea2017vehicle} manually annotate 677 taillight examples from the CompCars dataset \cite{yang2015large} to train their tail light segmentation model.}
\centering
\begin{tabular}{|C{3.5cm}|C{3cm}|C{8.5cm}|}
\hline
\textbf{Dataset Name} & \textbf{\# of Training Traffic Scene Images} & \textbf{Key Features} \\
\hline \hline
\href{https://cerv.aut.ac.nz/vehicle-lights-dataset/}{Vehicle Lights Dataset} \cite{rapson2019performance} & 1,232 & segmentation labels, data from different continents, variety of lighting and weather conditions \\
\hline
\href{http://mmlab.ie.cuhk.edu.hk/ datasets/comp_cars/index.html}{CompCars dataset} \cite{yang2015large} & 677 & images focused on head and taillights, variety of vehicle orientations\\
\hline
\href{https://cvrr.ucsd.edu/vehicle-lights-dataset}{LISA Vehicle Lights Dataset: Extension of ApolloCar3Dc} &  \textbf{3,634} & corner labels of keypoints, variety of lighting, vehicle light shapes, and orientations, vehicle-only context images with vehicle light centered in image \\
\hline
\end{tabular}
\label{table:datasetcomp}
\end{table*}

\begin{figure}[t]
  \centering
   \includegraphics[width=1\linewidth]{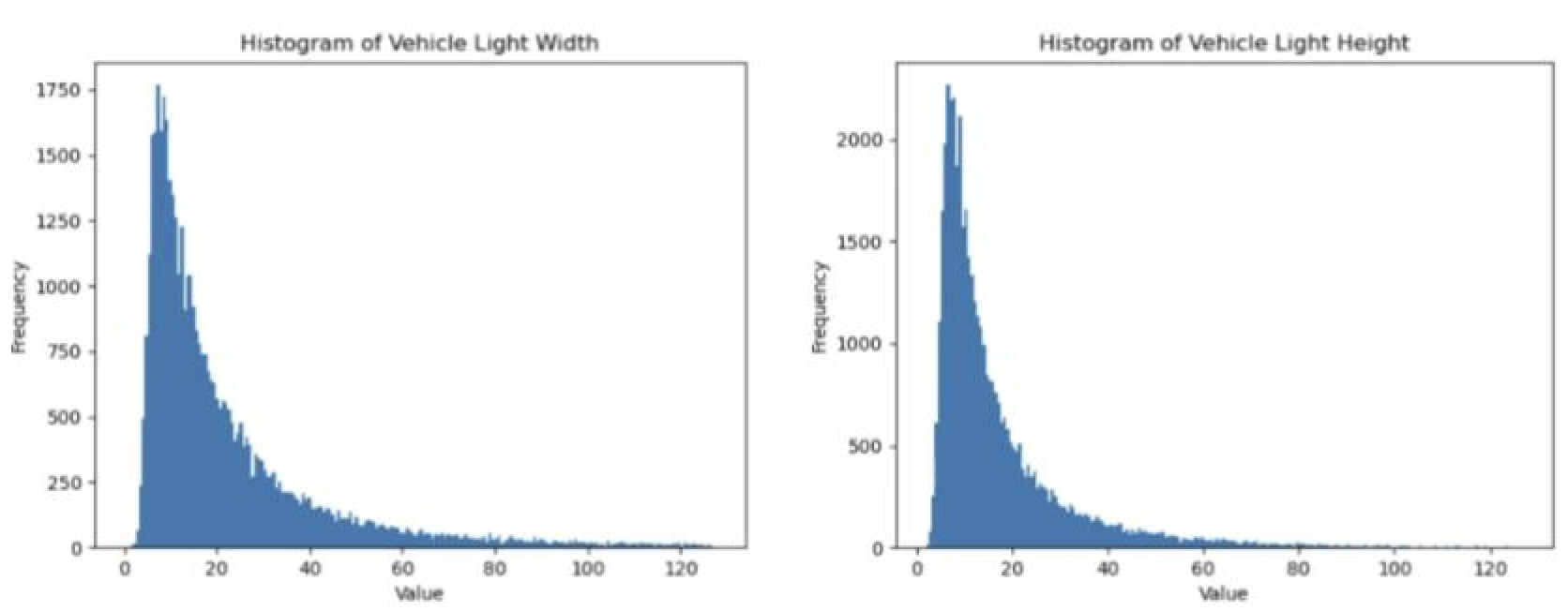}
   \caption{Histograms for the vehicle light heights and widths in the dataset we curate using the Apollo-Car3D dataset for vehicle light corner regression. The majority of the heights and widths of vehicle light tend to be in between 10-20 pixels.}
   \label{histogram}
\end{figure}

\begin{figure*}[hbt!]
  \centering
   \includegraphics[width=0.7\linewidth ]{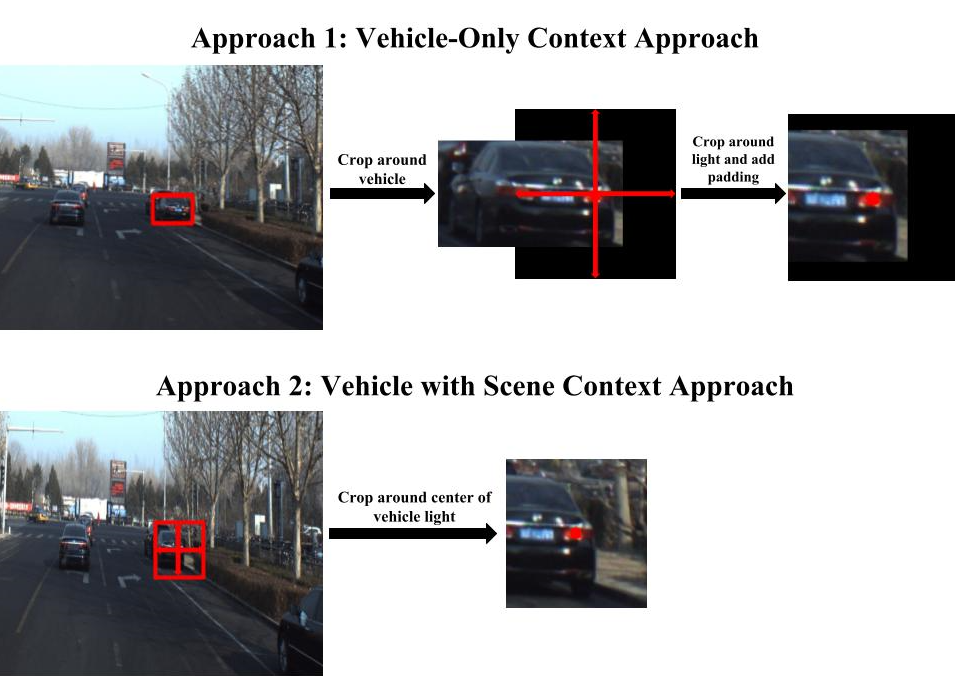}

   \caption{The two processes to create the cropped image around the center of the vehicle light. In the first approach, we crop around the vehicle and then use this image to crop around the center of the vehicle light of interest. For the second approach, we crop around the center of the vehicle light using the full image.}
   \label{fig2}
\end{figure*}

We filter the ApolloCar3D dataset \cite{song2019apollocar3d} to only use the keypoint annotations of any visible front and tail lights, which provide us with the $(x,y)$ coordinates in respect to the the full traffic scene image for the four corners of a vehicle light and its center. 

To associate lights to a vehicle, for each vehicle light annotation, we first crop the full traffic image to only focus on the vehicle of interest using the corresponding bounding box of the vehicle. With this cropped image of the vehicle, we create another $128\times 128$ cropped image that centers the center-coordinate of the taillight. To do this, we first calculate the translated coordinates $(x',y')$ of the vehicle light center so that they are in respect to the cropped vehicle image as follows: 
\begin{equation}
\begin{aligned}
    &x' = x - x_{bbox}\\
    &y' = y - y_{bbox}
\end{aligned}
\end{equation}
$(x,y)$ define the coordinates of the center of vehicle light from the traffic scene image and $(x_{bbox}, y_{bbox})$ represent the upper-left coordinates of the vehicle in the traffic scene image. Using this coordinate, we find the upper left and bottom right coordinates of the vehicle light cropped image by subtracting and adding 64 respectively to $(x',y')$. Once we know the upper left and bottom right coordinates, we crop the vehicle image around this region. In case these crops exceed the cropped vehicle image boundaries, we add black padding on the top, bottom, left, and right if necessary. This highlights the intended use of this dataset; if a light can be found using a center-based approach, but the full visual field of the light is desired (e.g. analysis of the light state to infer a signal), then it is desirable to generate an image in this form, because placing the center of the vehicle light in the middle of the image may simplify the learning process for detection models to locate outside edges or corners, as it will know consistently that the vehicle light of interest is located at the center of the image. We define this approach of first cropping around the vehicle and secondly the vehicle light as the ``Vehicle-Only Context Approach". We also use a second approach we coin the "Vehicle with Scene Context Approach" that includes further traffic context; in this second approach, we still place the center of the vehicle light in the middle of the image, but include image content from the entire traffic scene rather than only the cropped vehicle region. This approach leads to less padded pixels and the possibility of extraneous lights included in the sampled image. Figure \ref{fig2} shows an example of each approach. 

We annotate each light with ``corner" targets. To create the targets, we compute the difference of the $x$ and $y$ coordinates between the vehicle light corner and the vehicle light center coordinate $(x_{center}, y_{center})$. This computation gives us a list 8 offsets: $[x'_1, y'_1, x'_2, y'_2,  x'_3, y'_3,  x'_4, y'_4]$, where $(x'_1, y'_1)$, $(x'_2, y'_2)$, $(x'_3, y'_3)$, $(x'_4, y'_4)$ represent the offset from the center coordinate of the vehicle light to the upper left, upper right, bottom left, and bottom right corners respectively. We normalize these values from [-1, 1], where a value of -1 corresponds to a corner being 64 pixels to the left/above the light center and a value of 1 corresponds to a corner being 64 pixels to the right/below the right center. In some cases, a corner may not be visible in an image. If so, then we set $(x'_i, y'_i) = (0,0)$, where $(x'_i, y'_i)$ are the x and y pixel distances from the $i$th corner to the center of the vehicle light, a signal that such a corner should be ignored in training. 

From this process, we collect 44,784 total cropped images around a vehicle light. In particular, there are 9,578 cropped images of the left-front light, 17,794 left-rear light images, 13,677 right-rear light images, and 4,452 right-front light images. Furthermore, data augmentation has improved performance in many autonomous driving tasks such as vehicle detection \cite{pokrywka2022yolo} and can increase the vehicle light dataset size. This propelled us to curate an extension to the dataset that horizontally reflects each cropped vehicle light image and corresponding regression labels, doubling our dataset size to have 89,568 examples. We also introduce a separate section of the dataset that takes the segmentation light labels from Rapson et al. \cite{Rapson2018a} and use connected component labeling to convert the segmentation labels into bounding boxes around the vehicle light. Since no vehicle information is provided in this dataset, we use the Vehicle with Scene Context Approach to crop each of the bounding box vehicle light annotations. This provides us with an additional 2,606 vehicle light images to add to the LISA Vehicle Lights Dataset that are from different dataset sources. Figure \ref{histogram} shows a histogram of the heights and widths of the vehicle lights in this dataset. do, .Table \ref{table:datasetcomp} highlights our curated dataset in comparison to other public datasets used for vehicle light detection. This specialized version of the ApolloCar3D dataset we use has nearly three times the amount of traffic scene images in comparison to other public vehicle light datasets, providing an abundance of vehicle light examples and a variety of lighting conditions, vehicle light shapes, and orientations.

\subsection{Determining Vehicle Light Visibility}

\begin{figure}
    \centering
    \begin{subfigure}
      \centering
      \includegraphics[width=.45\linewidth]{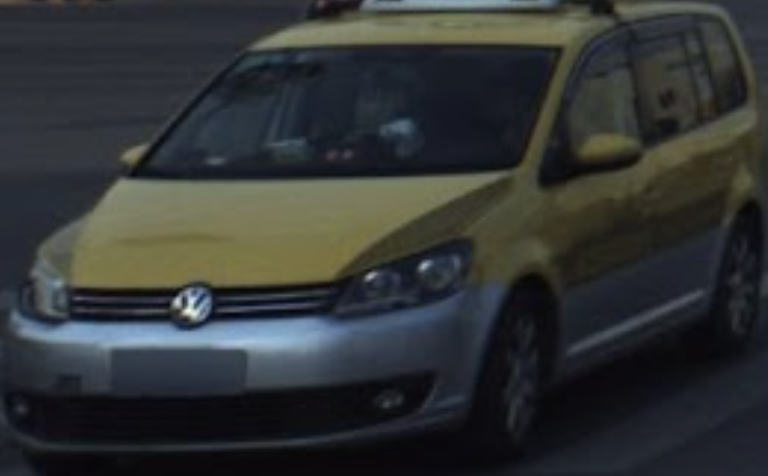}  
    \end{subfigure}
    \begin{subfigure}
      \centering
      \includegraphics[width=.45\linewidth]{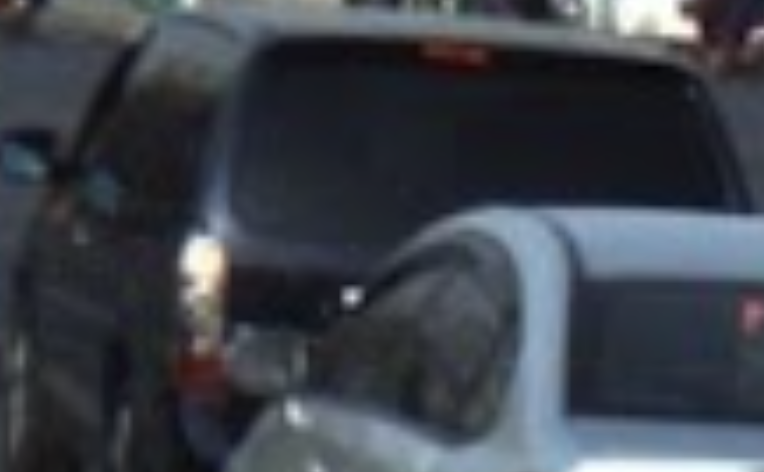}  
    \end{subfigure}
    
    \begin{subfigure}
      \centering
      \includegraphics[width=.45\linewidth]{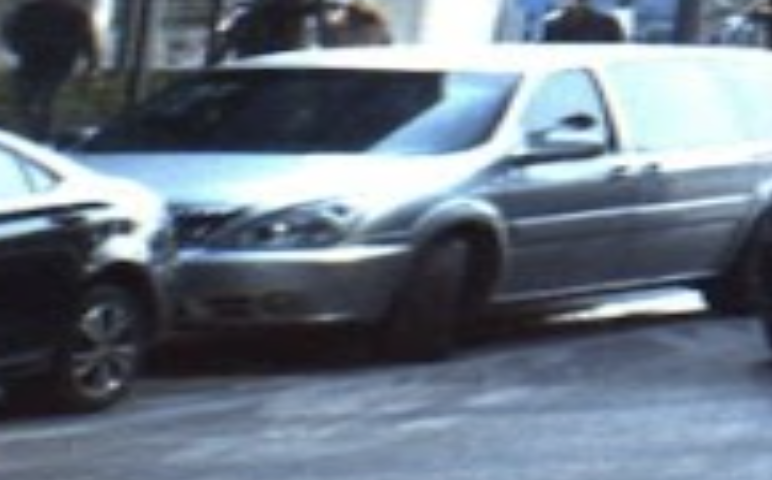}  
    \end{subfigure}
    \begin{subfigure}
      \centering
      \includegraphics[width=.45\linewidth]{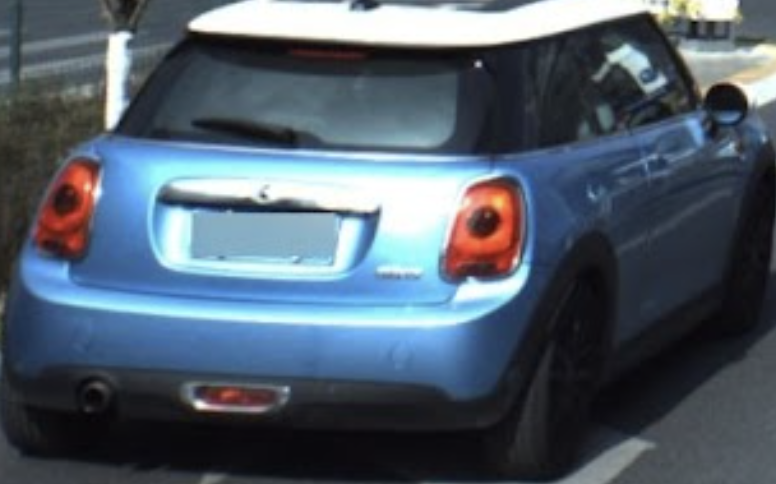}  
    \end{subfigure}
    \caption{Example orientations and situations of vehicles that affect the visibility of a vehicle light. In the top left image, only the front lights are visible. In the top right image, the rear left light is visible. In the bottom left image, the front left light is visible, and in the bottom right image, both rear lights are visible. When detecting the vehicle lights in the situations shown, it is necessary to determine the which lights are visible so that neural network output can be interpreted accurately.}
    \label{fig:lights_visibility}
\end{figure}

\begin{table}[hbt!]
\centering
\caption{Visibility performance for each of the single vehicle light models.}
\begin{tabular}{|c|C{2cm}|c|}
\hline
\textbf{Vehicle Light} & \textbf{Dataset Size (\# of Images)} & \textbf{\% Accuracy} \\
\hline \hline
Front-Left & 1,872 & 92.02 \\
\hline
Rear-Left & 1,947 & \textbf{96.41} \\
\hline
Front-Right & 2,826 & 91.0 \\
\hline
Rear-Right &  \textbf{4,859} & 90.76 \\
\hline
\end{tabular}
\label{table:visibility}
\end{table}

Because current models operate on a fixed output size, in order to completely perform vehicle light detection, a system must be able to determine which vehicle lights are visible and possible to be localized, so there are no false vehicle light detections for occluded or otherwise non-visible lights. We construct a group of convolutional neural networks which predict the visibility (but not location) of each of the four light assemblies. To train this model, we annotate the ApolloCar3D dataset with a binary tag per light assembly, indicating whether the particular light is visible. Vehicle detections which are too small or low quality are excluded. With the curated visibility vehicle light dataset, we then train 4 separate models, one for each type of vehicle light, each taking as input a cropped image of a vehicle from an upstream vehicle detector as input and predicting whether the vehicle light is visible or non-visible. These vehicle light visibility networks are built from a ResNet-34 baseline \cite{he2016deep} with a modified classifier layer to have one neuron output to represent the probability a vehicle light of interest is visible. We use binary cross entropy loss to train each visibility network and use an Adam optimizer \cite{kingma2014adam} to update the model weights. As Figure \ref{table:visibility} highlights, the visibility networks for each vehicle light are all over 90\% accuracy, with the rear left light model achieving the best visibility accuracy. The link to the code to train the vehicle light visibility models can be found \href{https://github.com/akshaygopalkr/RearLicenseVisbilityClassifier}{here}.

\section{Concluding Remarks}

\subsection{On the Order of Detection in Cascaded Models}
For the three vehicle safety tasks outlined in the introduction, and for related extensions, it is insufficient for a system to only detect vehicle lights. It is crucial that the system detect and \textit{associate} the lights with the correct vehicle. Returning to our opening discussion in framing the concept of object detection, for this task, the real requirement is the detection of a light assembly as related and joined to a particular vehicle instance. 

This introduces a selection of options when modeling this task. The detection of the vehicle and associated lights can be done simultaneously; the detection of the vehicle can be handled first, and the lights detected from this vehicle information; or, the lights can be detected first, then associated with a vehicle. 

As illustrated in Figure \ref{fig:order}, selection of this option is highly dependent on the task at hand, and the data available for training a model should support (rather than constrain) this selection, motivating our introduction of the LISA Vehicle Lights Dataset and Light Visibility Model. 

For example, if addressing the task of detecting surrounding vehicles in low-light (nighttime) conditions for obstacle avoidance, a system must first detect the lights, and using the pair of lights infer that there is an associated vehicle. On the other hand, if driving during the daytime and a planner must predict the path of a leading vehicle which is already detected, then the system may want to look to the signals, requiring secondary detection of the vehicle lights in a cascaded fashion, or use of a model like CenterNet which perform detection of objects and keypoints simultaneously. Again, in all scenarios, occlusion of a light or out-of-frame positioning is possible, so it is necessary for the system to have some level of understanding during training and inference so that all training constraints and object estimates are reasonable to our known concept of vehicle layout. 

\subsection{Extensions and Future Research}

The temporal nature of information from vehicle lights also suggests a future extension of vehicle light datasets: a series of annotated frames containing vehicle lights over time. This would assist models in learning a variety of tasks, from tracking and associating lights to the correct vehicles (and their symmetric twin, if visible), as well as analysis of  dynamic lighting patterns and interpretation of vehicle intention and trajectory. 

Framing the problem of vehicle light detection and association is a non-standard task because it requires consideration of what information is actually available in the driving environment. By offering a dataset and visibility model which provides vehicle light information in direct association with detected vehicles, this research enables further development in explainable, cascaded models which extract image fields containing individual vehicle lights, facilitating downstream analysis of the visual signals and cues offered to observing drivers. 

\bibliographystyle{ieee_fullname}
\bibliography{refs}

\end{document}